\documentclass[10pt,twocolumn,letterpaper]{article}

\usepackage{iccv}
\usepackage{times}
\usepackage{epsfig}
\usepackage{graphicx}
\usepackage{amsmath}
\usepackage{amssymb}

\usepackage{gensymb}
\usepackage{cite}
\usepackage{algorithmic}
\usepackage{textcomp}
\usepackage{scrextend}
\usepackage{changepage}
\usepackage[margin=0.75in]{geometry}
\usepackage{multirow}
\usepackage{amsfonts}
\usepackage[table]{xcolor}
\usepackage{booktabs}
\usepackage{mathtools}

\usepackage[breaklinks=true,bookmarks=false]{hyperref}

\iccvfinalcopy 

\def\iccvPaperID{7429} 

\ificcvfinal\fi

\begin{document}

\title{LSG-CPD: Coherent Point Drift with Local Surface Geometry \\ for Point Cloud Registration}

\author{Weixiao Liu $^{1, 2}$ 
\quad Hongtao Wu $^{1, 2}$~~~Gregory S. Chirikjian$^{1,2}$\footnotemark[1] \\
$^1$National University of Singapore, Singapore \quad
$^2$Johns Hopkins University, USA\\

{\tt\small \{mpewxl, mpehtw, mpegre\}@nus.edu.sg}
}

\maketitle

\renewcommand{\thefootnote}{\fnsymbol{footnote}}
\footnotetext[1]{Corresponding author}

\ificcvfinal\thispagestyle{empty}\fi

\begin{abstract}
Probabilistic point cloud registration methods are becoming more popular because of their robustness. However, unlike point-to-plane variants of iterative closest point (ICP)  which incorporate local surface geometric information such as surface normals, most probabilistic methods (e.g., coherent point drift (CPD)) ignore such information and build Gaussian mixture models (GMMs) with isotropic Gaussian covariances. This results in sphere-like GMM components which only penalize the point-to-point distance between the two point clouds. In this paper, we propose a novel method called CPD with Local Surface Geometry (LSG-CPD) for rigid point cloud registration. Our method adaptively adds different levels of point-to-plane penalization on top of the point-to-point penalization based on the flatness of the local surface. This results in GMM components with anisotropic covariances. We formulate point cloud registration as a maximum likelihood estimation (MLE) problem and solve it with the Expectation-Maximization (EM) algorithm. In the E step, we demonstrate that the computation can be recast into simple matrix manipulations and efficiently computed on a GPU. In the M step, we perform an unconstrained optimization on a matrix Lie group to efficiently update the rigid transformation of the registration. The proposed method outperforms state-of-the-art algorithms in terms of accuracy and robustness on various datasets captured with range scanners, RGBD cameras, and LiDARs. Also, it is significantly faster than modern implementations of CPD. The source code is available at \url{https://github.com/ChirikjianLab/LSG-CPD.git}.
\end{abstract}

\section{Introduction}

\begin{figure}
    \centering
    \includegraphics[width=0.86\columnwidth]{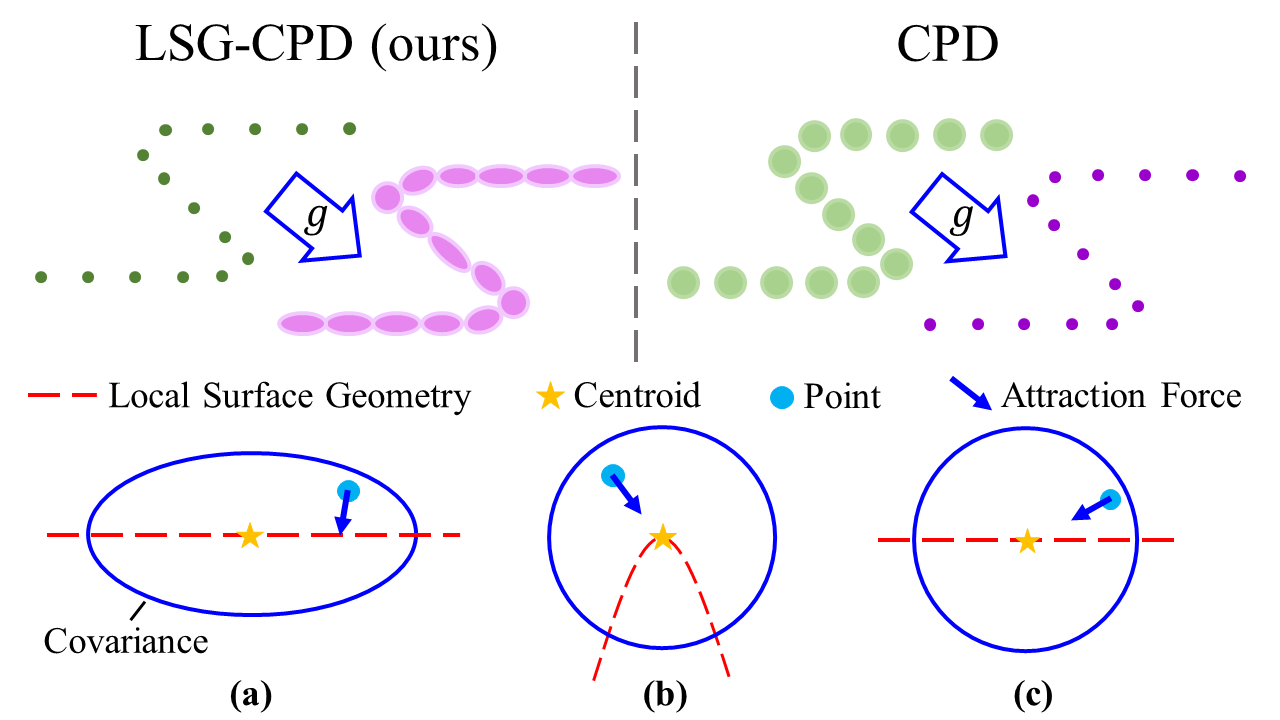} 
    \caption{CPD with Local Surface Geometry (LSG-CPD). Green denotes the source set; purple denotes the target set. Unlike CPD \cite{myronenko2010point} which constructs a sphere-like GMM component (c) for each point in the source set, the proposed method constructs GMM components on the target set. In the proposed method, plane-like GMM components (a) are created for points with small surface variation \cite{surface_variance}, \textit{i.e.}, flat local surface; sphere-like GMM components (b) are created for points with large surface variation.}
    \label{fig:concept}
    \vspace{-0.5cm}
\end{figure}

\label{sec: intro}
Point cloud registration plays an important role in various computer vision tasks \cite{nuchter20076d, whelan2015elasticfusion, chen1992object, schmidt2014dart}.
It is a task of aligning two point clouds and estimating the relative transformation between them.
When the correspondences between the points are known, methods have been proposed to solve it efficiently \cite{briales2017convex, yang2020teaser}.
Our paper focuses on the correspondence-free registration.
The most popular method, ICP \cite{besl1992ICP}, iteratively computes the alignment from point-to-point correspondence obtained via the closest point.
Various point-to-plane ICP variants \cite{chen1992object, Segal-RSS-09, chetverikov2002trimmed, NICP} make use of surface normals and take advantage of the tendency that most point cloud data are locally planar.
However, ICP and its variants require good initialization and are susceptible to outliers, noise, and missing points.

Probabilistic methods formulate point cloud registration as a probability density estimation problem \cite{myronenko2010point, gao2019filterreg, jian2010robust, stoyanov2012fast}.
In particular, methods such as CPD \cite{myronenko2010point} and FilterReg \cite{gao2019filterreg} describe the geometry of the first point cloud via a probability distribution over Euclidean space with a GMM.
The GMM is constructed by placing a Gaussian component with isotropic covariance at every point in the set.
The second set is fitted to it via MLE.
The introduction of the outlier component in the GMM and the sphere-like isotropic covariance makes these methods robust.
However, unlike point-to-plane ICP variants \cite{chen1992object, chetverikov2002trimmed, Segal-RSS-09, NICP} which take advantage of the local planar geometry in point cloud data, these methods ignore such information.
While these methods are robust, the isotropic covariance does not faithfully capture the local planar geometry of the point clouds, leading to inaccuracy in registration.

In this paper, we propose a novel method, CPD with Local Surface Geometry (LSG-CPD), which takes into account local surface geometry: surface normal and surface variation.
Surface variation \cite{surface_variance} is closely related to curvature and surface flatness.
Intuitively, the flatter the local surface, the smaller the curvature, the smaller the surface variation.
We construct a GMM on the target set.
The source set is fitted to it via MLE (Fig. \ref{fig:concept}).
However, unlike previous methods \cite{myronenko2010point, gao2019filterreg} which restrict covariances to be isotropic, we adaptively impose different levels of point-to-plane penalization on top of the original point-to-point penalization, resulting in anisotropic covariances (Sec. \ref{sec: math}).
If the local surface is flat, \textit{i.e.}, the surface variation is small, a large point-to-plane penalization is added.
The GMM component will be plane-like and points will be mainly attracted towards the local plane instead of the centroid (Fig. \ref{fig:concept}(a)).
If the local surface has large curvature or the normal estimation is noisy, \textit{i.e.}, the surface variation is large, the point-to-plane penalization will be small (Fig. \ref{fig:concept}(b)).
In this case, the GMM component will be sphere-like and robustness is maintained -- the registration will not be misled by those ineffectively estimated surface normals.
In addition, we propose a method to estimate the weight placed on the outlier component of the GMM (Sec. \ref{sec: outlier}).
Furthermore, we introduce the idea of confidence filtering (CF) which allows flexibility in reallocating weights of GMM components and outlier weights based on measurement confidence of points (Sec. \ref{sec:confidence}).
The EM algorithm \cite{bishop2006pattern, DEMP1977} is used to solve the MLE problem.
To make our method efficient, we reformulate the computation in the E step into parallelizable matrix manipulations which can be efficiently computed on a GPU (Sec. \ref{sec:CorM}). 
In the M step, we propose an unconstrained optimization method on a matrix Lie group and find the optimum with the efficient Newton's method (Sec. \ref{sec: MUOpt}).

Our motivation is to make the GMM-based description of the point cloud geometry richer and more faithful to the ``true" probabilistic distribution by incorporating local surface geometry, afforded by the surface normal and variation, with the original point position.
The optimum of the objective function in the EM process will be closer to the ``true" transformation compared to that using the isotropic covariances.
Thus, our method is able to improve the registration accuracy while maintaining the robustness of probabilistic methods.
Results show that the proposed method outperforms state-of-the-art algorithms in terms of accuracy and robustness on a wide variety of dataset captured by range scanners, RGBD cameras, and LiDARs.

\section{Related Works}

Point cloud registration has been extensively studied in the literature.
In this review, we limit the scope to GMM-based probabilistic methods.
One popular formulation of GMM-based methods builds a GMM distribution on one point set and softly assigns correspondence between each pair of points in the two sets probabilistically \cite{rangarajan1997robust, luo2003unified, mcneill2006probabilistic, granger2002multi, myronenko2010point, gao2019filterreg, horaud2010rigid, eckart2018hgmr, min2019robust, ravikumar2017generalised}.
Our method falls into this category.
The most related works are CPD \cite{myronenko2010point} and FilterReg \cite{gao2019filterreg}.
CPD \cite{myronenko2010point} derives the EM procedure and proposes a close-form solution in the M step.
Recently, FilterReg \cite{gao2019filterreg} introduces Gaussian filter and twist parameterization to improve the efficiency of the CPD framework.
Point-to-point FilterReg \cite{gao2019filterreg} is similar to CPD \cite{myronenko2010point}.
However, both methods ignore local surface geometry and place an isotropic covariance around every point in the first point cloud to construct the GMM.
Point-to-plane FilterReg \cite{gao2019filterreg} tries to account for the surface normal by substituting the point-to-point penalization with point-to-plane penalization in the M step.
This leads to various problems including rank deficiency of the covariance and inconsistency between the probabilistic model in the E and M step.
A more technical analysis can be found in  Sec. \ref{sec: math}.
ECMPR \cite{horaud2010rigid} and \cite{min2019_nonrigid} update the whole covariance matrix in the M step.
Although it allows general anisotropic covariances, no local geometric information is considered.
GMM-Tree \cite{eckart2018hgmr} partitions points into clusters and builds a hierarchy of Gaussian mixtures for the clusters.
The covariances, estimated directly by all the points in the cluster instead of local surface geometry, are anisotropic.
But it does not account for outliers in its probabilistic modeling, making it not robust to outliers and missing points.
In \cite{min2019robust, ravikumar2017generalised}, hybrid mixture models (\textit{e.g.}, Gaussian $+$ von-Mises-Fisher) are used to model position and orientation uncertainty.
The complicated mixture models introduce heavy computation.
In contrast, our method incorporates surface normals in a simple GMM adaptively, making it robust to ineffective surface normal estimation.

Another popular line of GMM-based methods build GMM probability distribution on both the source and the target sets \cite{jian2010robust, tabib2018manifold, stoyanov2012fast, campbell2015adaptive}.
Point cloud registration is formulated as a problem of minimizing the difference between the two distributions.
\cite{jian2010robust} uses KL-divergence to parameterize the distribution difference while \cite{tabib2018manifold} uses squared L2 norm.
To improve the efficiency, voxelization \cite{stoyanov2012fast} and Support Vector Machine \cite{campbell2015adaptive} are used to represent the probability representation in a compact and efficient way.

\section{Methods}

\subsection{Formulation of Probabilistic Model}
\label{sec: math}
The alignment of two point clouds is formulated as an MLE problem. 
We fix one point cloud (target set) and align the other one (source set) to it.
Unlike CPD \cite{myronenko2010point} which places the GMM on the source set, the GMM in our method is built on the target set, which is similar to FilterReg\cite{gao2019filterreg} (Fig. \ref{fig:concept}). 
Rigid transformation is applied on the source set and the transformed source set is regarded as the observation of the GMM. 
The goal is to find the optimal transformation which makes the observation best fit the GMM.
Through out the paper, we adopt the following notation: 
\begin{itemize}
\vspace{-0.2cm} 
\setlength\itemsep{-0.4em} 
\item $N$ - number of points in the source set
\item $M$ - number of points in the target set
\item $\mathbf{X}\in\mathbb{R}_{N\times 3}$ - source set where the $n$-th row $\mathbf{x}_n^T$ is the position of the $n$-th point in the set
\item $\mathbf{Y}\in\mathbb{R}_{M\times 3}$ - target set where the $m$-th row $\mathbf{y}_m^T$ is the position of the $m$-th point in the set
\item $\mathbf{N}\in\mathbb{R}_{M\times 3}$ - surface normal matrix  where the $m$-th row $\mathbf{n}_m^T$ is the unit normal vector of the corresponding $m$-th point in $\mathbf{Y}$,
\item $\mathbf{I}$ - identity matrix
\vspace{-0.2cm} 
\end{itemize}
Following \cite{myronenko2010point, gao2019filterreg, eckart2018hgmr}, the probability density function of the GMM is given as:
\begin{equation}
    p(\mathbf{x}_n)=w_{n}p_{o}(\mathbf{x}_n)+(1-w_{n})\sum_{m=1}^{M}\pi(m)p(\mathbf{x}_n|m)
\label{eqn:1}
\end{equation}
where $\pi(m)$ is the prior probability of $\mathbf{x}_n$ being assigned to the $m$-th component; $p(\mathbf{x}_n|m)$ is the corresponding component density function.
Besides the $M$ Gaussian components centered at each point in the target set, an outlier component is added to the GMM to make the algorithm robust to outliers and missing points.
$w_{n}\in [0, 1)$ is the outlier weight of $\mathbf{x}_n$ and $p_{o}$ is the outlier probability density. 

In \cite{myronenko2010point, gao2019filterreg}, the outlier probability density $p_o(\mathbf{x}_n) = \frac{1}{N}$. 
As pointed out in \cite{hirose2020bayesian}, it does not satisfy a property of probability density functions: the integral of $p_{o}(\mathbf{x}_n)$ over $\mathbb{R}^{3}$ equals 1.
In this paper, we follow \cite{hirose2020bayesian} and define a working space $\mathbb{V} \subset \mathbb{R}^{3}$ whose volume equals $V$.
$\mathbb{V}$ represents a solid body which encapsulates the target set.
$p_{o}(\mathbf{x}_n) = 1/V$ if $\mathbf{x}_n \in \mathbb{V}$ and $p_{o}(\mathbf{x}_n) = 0$ otherwise.
In \cite{myronenko2010point, gao2019filterreg}, $w_n = w_0$ for $n=1, 2, ..., N$; $\pi(m) = \frac{1}{M}$ for $m=1, 2, ..., M$.
Both $w_n$ and $\pi(m)$ are uniform across all points in the source and target set, respectively.
In Sec. \ref{sec: outlier}, we introduce a method to estimate $w_0$ from the outlier ratio of the point cloud.
In Sec. \ref{sec:confidence}, we propose a method which allows flexibility to assign different $w_n$ and $\pi(m)$ to different $\mathbf{x}_n$ and $\mathbf{y}_m$ respectively according to their measurement confidence.

The key novelty of our method lies in our design of $p(\mathbf{x}_n|m)$ which incorporates both point-to-point and point-to-plane penalization into the probabilistic modeling.
Surface normals are easily obtainable features of point cloud data \cite{hoppe1992surface, surface_variance} and have been successfully utilized in many ICP variants \cite{chen1992object, Segal-RSS-09, 10.1007/978-3-319-10404-1_23, NICP} to improve the registration performance.
Point-to-plane extension of FilterReg \cite{gao2019filterreg} tries to extend the CPD \cite{myronenko2010point} framework by substituting the point-to-point distance in the original objective function with the point-to-plane distance in the M step without further modifying the E step.
This will lead to rank deficiency of the covariances, thus resulting in an ill-posed optimization in the M step.
More importantly, the underlying probabilistic model in the E and M step are inconsistent. 
Neither convergence nor optimality is guaranteed for the EM process \cite{wu1983EMConverge}.
In this paper, we propose a component density function as follows:
\begin{equation}
\begin{gathered}
p(\mathbf{x}_n|m) =  c_m\exp{\left( -\frac{1}{2}(\mathbf{x}_n-\mathbf{y}_m)^T \mathbf{\Sigma}_m^{-1} (\mathbf{x}_n-\mathbf{y}_m) \right)} \\
\mathbf{\Sigma}_m^{-1} = \frac{1}{\sigma^2}(\alpha_m \mathbf{n}_m\mathbf{n}_m^T + \mathbf{I})
\end{gathered}
\label{eqn:2}
\end{equation}
$c_m$ is a normalizing constant.
$\mathbf{\Sigma}_m$ is a full-rank anisotropic covariance matrix of which the inverse encodes information of the local surface geometry.
$\sigma^2$ is the covariance multiplier.
It is initialized autonomously similar to CPD \cite{myronenko2010point}.
The identity matrix $\mathbf{I}$ in $\mathbf{\Sigma}_m^{-1}$ penalizes the point-to-point distance between $\mathbf{x}_n$ and $\mathbf{y}_m$ similar to \cite{myronenko2010point, gao2019filterreg}.
The rank one projection matrix $\mathbf{n}_m\mathbf{n}_m^T$ penalizes the point-to-plane distance between $\mathbf{x}_n$ and the local plane defined by $\mathbf{n}_{m}$ at $\mathbf{y}_m$.
These two matrices mutually benefit each other.
$\mathbf{n}_m\mathbf{n}_m^T$ enhances the accuracy provided by the point-to-point penalization.
Meanwhile, $\mathbf{I}$ serves as a regularizer, pulling the problem away from ill-posedness. 
The penalization coefficient $\alpha_m$ is set adaptively to add different levels of point-to-plane penalization depending on how flat the local surface is. 
This is realized by evaluating the surface variation $\kappa_m$ \cite{surface_variance} within the neighborhood of $\mathbf{y}_m$. 
Surface variation $\kappa\in[0, \frac{1}{3}]$ \cite{surface_variance} is the ratio of the smallest eigenvalue to the trace of the covariance matrix of a local neighborhood (note that this covariance matrix is not $\mathbf{\Sigma}_{m}$).
It is computed via eigenanalysis.
It describes the degree of deviation of points from the tangent plane. 
Intuitively, the flatter the local surface, the smaller the curvature, the smaller the surface variation.
$\alpha_m$ is chosen as a modified logistic sigmoid function:
\begin{equation}
    \alpha_m = \frac{1 - \exp{(\lambda(3-\frac{1}{\kappa_m}))}}{1 + \exp{(\lambda(3-\frac{1}{\kappa_m}))}}\alpha_\mathrm{max}
\end{equation}
where $\alpha_\mathrm{max}$ is the upper bound; 
$\lambda$ is a coefficient governing the sensitivity of $\alpha_m$ to $\frac{1}{\kappa_m}$.
$\alpha_m\to 0$ when $\kappa_m\to \frac{1}{3}$ (large surface variation, non-flat local surface geometry);
$\alpha_m\to\alpha_{max}$ when $\kappa_m\to 0$ (small surface variation, flat local surface geometry). 
By making $\alpha_m$ adaptive to local surface geometry, our algorithm obtains several desirable properties: 
(1) large point-to-plane penalization is only applied to points around which the local surface can be well approximated by a plane;
(2) little directional preference is imposed when the curvature is large or effective normal estimation is unavailable, \textit{e.g.}, corrupted by noise.

We denote $g(\mathbf{x}_n)$ as the image of the point $\mathbf{x}_n$ transformed by the rigid transformation $g$.
Given the GMM and the observation, the negative log-likelihood function is:
\begin{equation}
    L(g, \sigma^2)=-\sum^{N}_{n=1}\log (p(g(\mathbf{x}_n)))
    \label{eqn:3}
\end{equation}
Our goal is to find the optimal rigid transformation which minimizes $L$.
However, directly solving this optimization problem is intractable.
Thus, the EM algorithm \cite{bishop2006pattern, DEMP1977} is applied to find the solution.
The correspondence between the points in $\mathbf{X}$ and $\mathbf{Y}$ is treated as a latent variable.

\textbf{E step:} We use the transformation $g_{old}$ from the previous iteration to compute the posterior correspondence probability between each pair of point in $\mathbf{X}$ and $\mathbf{Y}$ via Bayes' rule. The correspondence probability is stored in a matrix $\mathbf{P} \in \mathbb{R}_{M \times N}$ where the $(m, n)$-th entry $\mathbf{P}_{mn}$ represents the posterior probability of the point $\mathbf{x}_n$ being assigned to the $m$-th component:
\begin{equation}
   \mathbf{P}_{mn}=\frac{(1-w_n)\pi(m)p(g_{old}(\mathbf{x}_n)|m)}{w_n p_o + (1-w_n) \sum_{m}\pi(m)p(g_{old}(\mathbf{x}_n)|m)}
\label{eqn:4}
\end{equation}
The computation of $\mathbf{P}$ can be reformulated into efficient and parallelizable matrix manipulations (Sec. \ref{sec:CorM}). 

\textbf{M step:} The rigid transformation $g$ and covariance multiplier $\sigma^2$ are inferred by minimizing the following \cite{myronenko2010point, bishop2006pattern} (see Supplementary Material for details):
\begin{equation}
    Q(g,\sigma^2)=-\sum^{N}_{n=1}\sum^{M}_{m=1}\mathbf{P}_{mn}\log \left( \pi(m)
    p(g(\mathbf{x}_n)|m) \right)
\label{eqn:5}
\end{equation}
In the case where the covariance matrix $\mathbf{\Sigma}_m$ is isotropic, a closed form solution of Eq. \eqref{eqn:5} can be found \cite{myronenko2010point}. 
However, this is not the case when $\mathbf{\Sigma}_m$ is anisotropic\cite{horaud2010rigid}. 
Thus, we propose an unconstrained optimization method on a matrix Lie group to solve the optimization in the M step in Sec. \ref{sec: MUOpt}.

\subsection{Outlier Weight Estimation}
\label{sec: outlier}
In \cite{myronenko2010point, gao2019filterreg}, the uniform outlier weight $w_0\in [0, 1)$ is manually tuned by the user. 
However, $w_0$ is merely a hyperparameter which is different from the outlier ratio $\eta$ - the ratio of outliers to the number of points in the source set.
This makes it hard to choose $w_0$ in different tasks intuitively.
Here, we introduce a method to determine $w_0$ from $\eta$ using the fact that $w_0$ and $\eta$ are probabilistically related.
In this way, the user only needs to set the outlier ratio $\eta$, which is more intuitive than setting $w_0$.
When the point clouds are well aligned by the ``true" transformation $g^*$, the expectation of the outlier number $N_{outlier}$ should equal the true number of outliers $\eta N$:
\begin{equation}
    E(N_{outlier})=\sum_{n=1}^{N}P(M+1|g^*(\mathbf{x}_n))=\eta N
\label{expectation_outlier}
\end{equation}
where
\begin{equation*}
P(M+1|g^{*}(\mathbf{x}_n)) = \frac{\frac{w_0}{1-w_0}p_o}{\frac{w_0}{1-w_0}p_o+\sum_{m}\pi(m)p(g^*(\mathbf{x}_n)|m)}
\end{equation*}
is the posterior probability of the point $\mathbf{x}_n$ being an outlier.
$w_0$ can then be solved from Eq. \eqref{expectation_outlier}.
However, $g^*$ is not available \textit{a priori}. 
Thus, from Eq. \eqref{expectation_outlier}, we can only derive the upper bound of $w_0$, \textit{i.e.}, $w_0 \leq w_\mathrm{max} =\frac{\eta V \sum_{m}\pi(m)c_m}{(1-\eta) + \eta V  \sum_{m}\pi(m)c_m} \in [0, 1)$, by setting $p(g^*(\mathbf{x}_n)|m)$ as its upper bound $c_m$  (Eq. \eqref{eqn:2}).
To make the algorithm robust, generally a large $w_0$ is desired.
However, if $w_0 > w_\mathrm{max}$, probabilistically, more points are regarded as outliers than the truth.
The inlier density has the risk of being submerged by the outlier density.
And as $w_0\to 1$, the density of the GMM will be primarily dominated by the outlier component. 
The algorithm tends to get stuck at a local minimum.
Therefore, we set $w_0=w_\mathrm{max}$ in practice to strike a balance.
\subsection{Confidence Filtering}
\label{sec:confidence}
Most GMM-based methods \cite{gao2019filterreg, horaud2010rigid, jian2010robust, myronenko2010point, eckart2018hgmr, min2019robust} weigh all the GMM components uniformly with $\pi(m) = \frac{1}{M}$. 
In addition, the outlier weight $w_n=w_0$ is uniform across all points in the source set.
This setting is reasonable when the measurement error induced by the sensor is uniform within its range, \textit{e.g.}, range scanners.
And thus the measurement confidence of all the points is equal in general.
However, point clouds acquired from consumer-grade RGBD cameras are subject to bias and noise which grows with the increment of depth \cite{basso2018robust, depth_camera_error_model}.
The larger the depth of the point is, the less confident (or more uncertain) the measurement is.
The algorithm should pay more attention to the alignment of points measured with high confidence.
In this paper, we define $\phi(\mathbf{x})=e_\mathrm{min} / e(\mathbf{x})$ as the measurement confidence of a point $\mathbf{x}$, where 
$e(\mathbf{x})$ is the error model which provides the statistical measurement error of a point measured at $\mathbf{x}$ \cite{depth_camera_error_model};
$e_\mathrm{min}$ is the minimum measurement error within the sensor range.
Error analysis and modeling of different sensors can be found in \cite{depth_camera_error_model,horaud2016overview, christian2013survey}.
The confidence filtering (CF) modifies $\pi(m)$ and $w_n$ as:
\begin{equation}
\begin{gathered}
    \pi(m)=\frac{
    \phi(\mathbf{y}_m)}
    {\sum_{M} \phi(\mathbf{y}_m)},\quad
    w_n=1-(1-w_0)\phi(\mathbf{x}_n)
\end{gathered}
\end{equation}
The component prior probability $\pi(m) \in (0, 1)$ decreases as $\phi(\mathbf{y}_m)$ decreases.
The reallocation of $\pi(m)$ reduces the attraction force of a component generated by a point with less confidence and vice versa.
The outlier weight $w_n \in [w_0, 1]$ increases as $\phi(\mathbf{x}_n)$ decreases.
$w_0$ is the baseline outlier weight determined in Sec. \ref{sec: outlier}.
As the measurement of $\mathbf{x}_n$ becomes less confident, we would rather categorize it as an ``outlier" to lower its contribution to the registration.
Furthermore, points with confidence lower than a threshold can be truncated.
This not only improves the accuracy but also reduces the runtime as shown in Sec. \ref{sec:RGBD}.
Note that the uniformly weighing setting is a special case where $\phi(\mathbf{x}) = 1$ for any $\mathbf{x}$.
\subsection{E Step: Efficient Computation of Correspondence Matrix}
\label{sec:CorM}
The introduction of the correspondence matrix $\mathbf{P}$ provides GMM-based methods with an inherent advantage on robustness. 
However, computing $\mathbf{P}$ is the most computationally expensive part of the algorithm: the correspondence probability of every pair of points in $\mathbf{X}$ and $\mathbf{Y}$ needs to be computed. 
We reformulate this process into simple matrix manipulations which can be efficiently computed on a GPU. 

Plugging $w_n$ and $\pi(m)$ defined in the previous sections and Eq. \eqref{eqn:2} into Eq. \eqref{eqn:4}, the $(m, n)$-th entry of the correspondence matrix $\mathbf{P}$ becomes:
\begin{equation}
    \mathbf{P}_{mn}=\frac{\mathbf{K}_{mn}}{\sum_{m=1}^M \mathbf{K}_{mn}+\gamma}
\end{equation}
where $\gamma = (2\pi\sigma^2)^{3/2} / V$ and $\mathbf{K}_{mn}$ is the $(m, n)$-th entry of a matrix $\mathbf{K}\in \mathbb{R}_{M\times N}$:
\begin{equation*}
\mathbf{K}_{mn} = \frac{1-w_n}{w_n}\pi(m)\exp \left(-\frac{1}{2}\lVert g(\mathbf{x}_n)-\mathbf{y}_m\rVert_{\mathbf{\Sigma}_m^{-1}}^2 \right)
\end{equation*}
Exploiting the special formulation of $\mathbf{\Sigma}_m$, the computation of $\mathbf{K}$ can be recast into fundamental matrix manipulations:
\begin{equation}
    \mathbf{K} = \mathbf{C}\odot \underline{\exp} \left( -\frac{1}{2\sigma^2}(\mathbf{D}\mathbf{A}+\mathbf{B}) \right)
\label{eqn:12}
\end{equation}
in which $\odot$ is the element-wise product of matrices; $\underline{\exp}(\cdot)$ is the element-wise exponential; the $(m, n)$-th entry of $\mathbf{A}$
\begin{equation*}
\begin{gathered}
    \mathbf{A}_{mn} = ((\mathbf{NRX}^T+\mathbf{Nt1}^T-\mathbf{s1}^T)_{mn})\textsuperscript{2}\\
    \mathbf{B} = \mathbf{Q}+\mathbf{t}^T\mathbf{tI}+2(\mathbf{1t}^T\mathbf{RX}^T-\mathbf{YRX}^T-\mathbf{Yt1}^T)\\
    \mathbf{C} \in\mathbb{R}_{M\times N},\: \mathbf{C}_{mn} = \frac{1-w_n}{w_n}\pi(m)\\
    \mathbf{D} = diag(\alpha_1,\alpha_2,\dots,\alpha_m)
\end{gathered}
\end{equation*}
in which $\mathbf{1}$ is the column vector with all ones,
\begin{equation*}
    \mathbf{s}=
    \begin{bmatrix}
    \mathbf{y}_1^T\mathbf{n}_1 & \mathbf{y}_2^T\mathbf{n}_2 & \dots & \mathbf{y}_M^T\mathbf{n}_M
    \end{bmatrix}^T
\end{equation*}
\begin{equation*}
\mathbf{Q}\in\mathbb{R}_{M\times N}, \mathbf{Q}_{mn} = \lVert \mathbf{x}_n\rVert_2^2+\lVert \mathbf{y}_m\rVert_2^2
\end{equation*}
are terms independent of transformation $g=(\mathbf{R}, \mathbf{t})$, and thus can be pre-computed.

\subsection{M Step: Optimization On Matrix Lie Group}
\label{sec: MUOpt}
In the M step, our goal is to infer the optimal transformation and variance multiplier $\sigma^2$ by minimizing $Q$ given by Eq. \eqref{eqn:5}. 
We propose an unconstrained optimization method to solve this optimization problem. 
Instead of parameterizing the rigid transformation, our method optimizes on the matrix Lie group via Newton's method. 
The gradient and the Hessian matrix of a function on matrix Lie group are defined with respect to the basis in Lie algebra \cite{stochastic}. 
We denote $\mathbf{E}_1, \mathbf{E}_2,..., \mathbf{E}_6$ as the basis matrices of $se(3)$ Lie algebra.
The gradient $\nabla Q = \begin{bmatrix} E_1^{r}Q & E_2^{r}Q & \dots & E_6^{r}Q \end{bmatrix}^T \in \mathbb{R}^{6}$.
The Hessian $\mathbf{H} \in \mathbb{R}_{6\times6}$ is asymmetric in which the $(i, j)$-th entry $\mathbf{H}_{ij}=E_i^rE_j^{r}Q$.
$E_i^r$ is the right derivative operators with respect to the $i$-th basis matrix $\mathbf{E}_i$. 
Both the gradient and the Hessian can be derived in closed-form:
\begin{equation}
\begin{aligned}
    E_i^r Q & \doteq \frac{d}{dt}Q(g\circ\exp(t\mathbf{E_i}))|_{t=0} \\
    & = 2\sum_{m=1}^{M}\sum_{n=1}^{N}\mathbf{P}_{mn}(\widetilde{g}\widetilde{\mathbf{x}}_n-\widetilde{\mathbf{y}}_m)^T\widetilde{\mathbf{\Sigma}}^{-1}_{m}\widetilde{g}\mathbf{E}_i\widetilde{\mathbf{x}}_n
\end{aligned}
\end{equation}

\begin{equation}
\begin{aligned}
    E_i^rE_j^{r}Q= &2\sum_{m=1}^M\sum_{n=1}^N
    \mathbf{P}_{mn}(\widetilde{\mathbf{x}}_n^T\mathbf{E}_j^T\widetilde{g}^T\widetilde{\mathbf{\Sigma}}^{-1}_{m}\widetilde{g}\mathbf{E}_i\widetilde{\mathbf{x}}_n\\
    &+(\widetilde{g}\widetilde{\mathbf{x}}_n-\widetilde{\mathbf{y}}_m)^T\widetilde{\mathbf{\Sigma}}^{-1}_{m}\widetilde{g}\mathbf{E}_j\mathbf{E}_i\widetilde{\mathbf{x}}_n)
\end{aligned}
\label{eqn:18}
\end{equation}
$\widetilde{g}$, $\widetilde{\mathbf{x}}_n$, and $\widetilde{\mathbf{y}}_m$ are the homogeneous form of $g$, $\mathbf{x}_n$, and $\mathbf{y}_m$, respectively. $\widetilde{\mathbf{\Sigma}}_m^{-1} = \mathbf{\Sigma}_m^{-1}\oplus0$ is the augmented inverse covariance matrix where $\oplus$ denotes the direct sum.
Detailed derivations can be found in the Supplementary Material.
It is worth noting that this method can be easily extended to other types of registration of which the transformation can be represented by a matrix Lie group, \textit{e.g.}, affine transformation, by replacing the basis matrices in Lie algebra.
The transformation is updated by Newton's method until convergence:
\begin{equation}
    \widetilde{g}_{i+1}=\widetilde{g}_{i}\circ \exp{(\frac{1}{2}{(\mathbf{H}+\mathbf{H}^T)^{-1}\nabla Q)}^{\wedge}}
\end{equation}
where $\wedge$ is the operation which maps the vector to the corresponding matrix Lie algebra \cite{murray1994mathematical}. 
After $g$ gets updated, $\sigma^2$ is inferred by setting the corresponding partial derivative of Eq. \eqref{eqn:5} to zero.
Unlike ECMPR\cite{horaud2010rigid} in which the whole covariance matrix is updated, only the covariance multiplier $\sigma^2$ gets updated in our approach.
As a result, while the volume of the covariance $\mathbf{\Sigma}_{m}$ is updated in every iteration, the local geometric structure is preserved.

\section{Results}
We conduct a wide variety of experiments on public 3D point cloud datasets, including both object and environment dataset, to show the robustness and accuracy of our method. 
We implement our method on Matlab.
We evaluate on a computer running Intel Core i9-9900K with an NVIDIA GeForce RTX 2080Ti GPU.
In all the experiments, the code of the baseline methods are either provided directly by the authors or taken from the popular open source library with various performance optimization.
The parameters for all the baseline methods are either provided by the authors or the softwares or well-tuned by ourselves if not provided.
CF is only used in the experiment on the RGBD dataset (Sec. \ref{sec:RGBD}).
The error model $e(\mathbf{x})$ of the sensor (Asus Xtion Pro), which is a quadratic function of the depth of $\mathbf{x}$, is sourced from \cite{depth_camera_error_model}.
More details on the experiment settings can be found in the Supplementary Material.

\subsection{Registration on Data with Outliers and Noise}
\label{sec: Outlier and Noise}
We follow \cite{myronenko2010point, gao2019filterreg} and test on laser range data by adding outliers and noise, respectively.
For both experiments, we use the Bunny dataset in Stanford 3D scanning repository \cite{curless1996volumetric}.
We downsample the original point cloud to about 3500 points and rotate it about a random axis by 50 degrees.
We compare with 4 baseline method: CPD \cite{myronenko2010point}
, FilterReg \cite{gao2019filterreg}, ECMPR \cite{horaud2010rigid}, and TrICP \cite{chetverikov2002trimmed}.
CPD and FilterReg represent GMM-based algorithm with isotropic covariances; ECMPR represents GMM-based algorithm with anisotropic covariances; TrICP, a widely used robust ICP variant, represents the ICP methods.
Following \cite{gao2019filterreg}, we use the corresponding point-to-point distance to measure the registration error:
\begin{equation}
    \textrm{error}(g) = \frac{1}{N}\sum^{N}_{i=1}\lVert g(\mathbf{x}_{i}) - g_{gt}(\mathbf{x}_{i}) \rVert_{2}
\label{eqn: multiview error}
\end{equation}
where $\mathbf{x}_{i}$ is the i-th point in the source set; 
$g$ and $g_{gt}$ are the transformations of the registration and the ground truth, respectively.
All methods are exhaustively run with 100 iterations to ensure convergence.
All the results are statistical: errors are averaged values of 30 independent runs.

\begin{figure} [!htp]
    \centering
    \includegraphics[width=1\columnwidth]{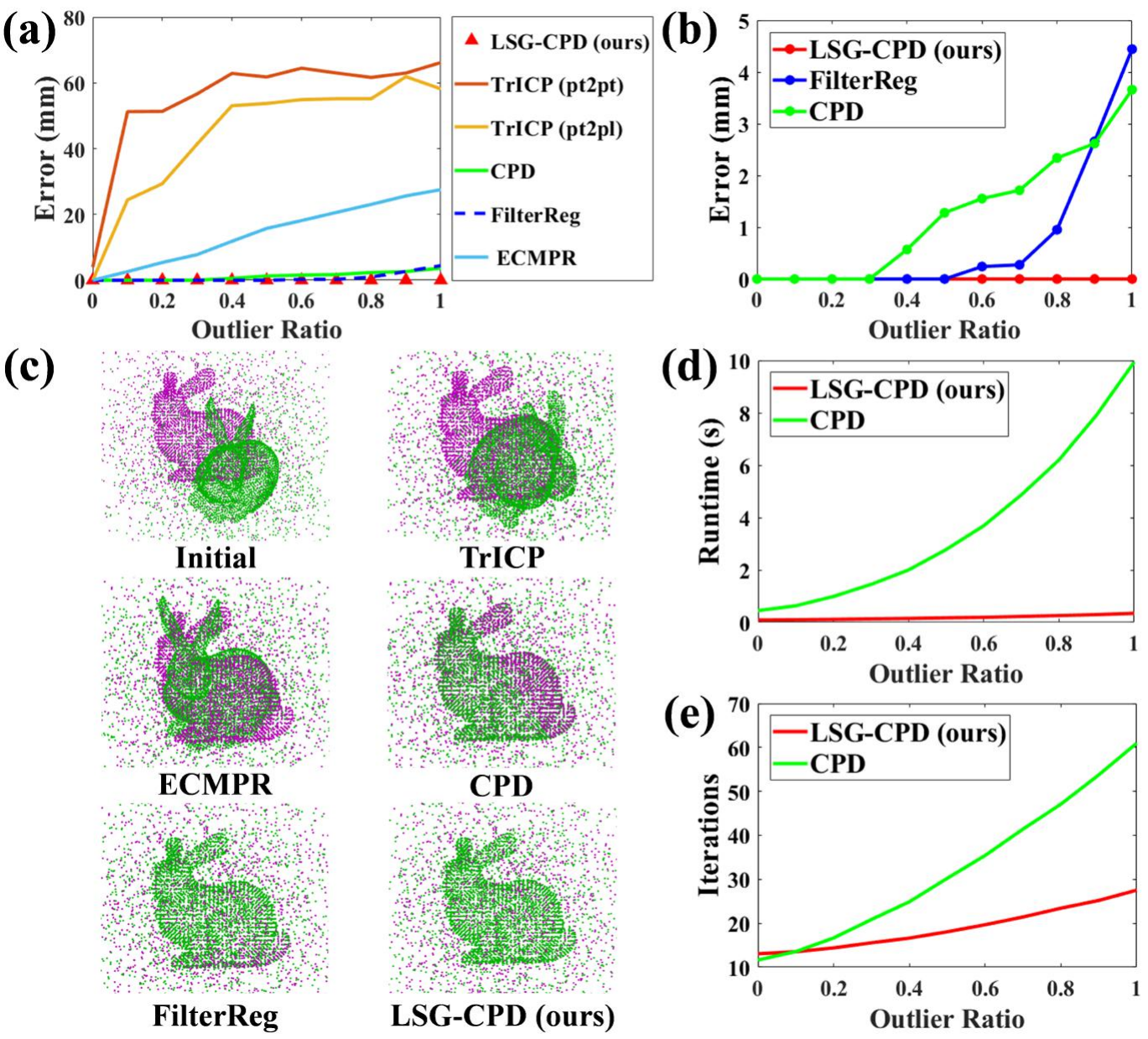} 
    \caption{Outlier results. (a) Errors (Eq. \eqref{eqn: multiview error}) of different methods. (b) Enlarged image showing errors of CPD, FilterReg, and LSG-CPD. (c) Initialization and results of different methods on data with 0.5 outlier ratio. TrICP point-to-plane result is shown as the result for TrICP. The images are cropped for better visualization. (d) Average runtime of CPD and LSG-CPD on data with different outlier ratios. (e) Averaged iterations of CPD and LSG-CPD on data with different outlier ratios.}
    \label{fig:outlier}
    \vspace{-0.2cm}
\end{figure}
\begin{figure} [!h]
    \centering
    \includegraphics[width=1\columnwidth]{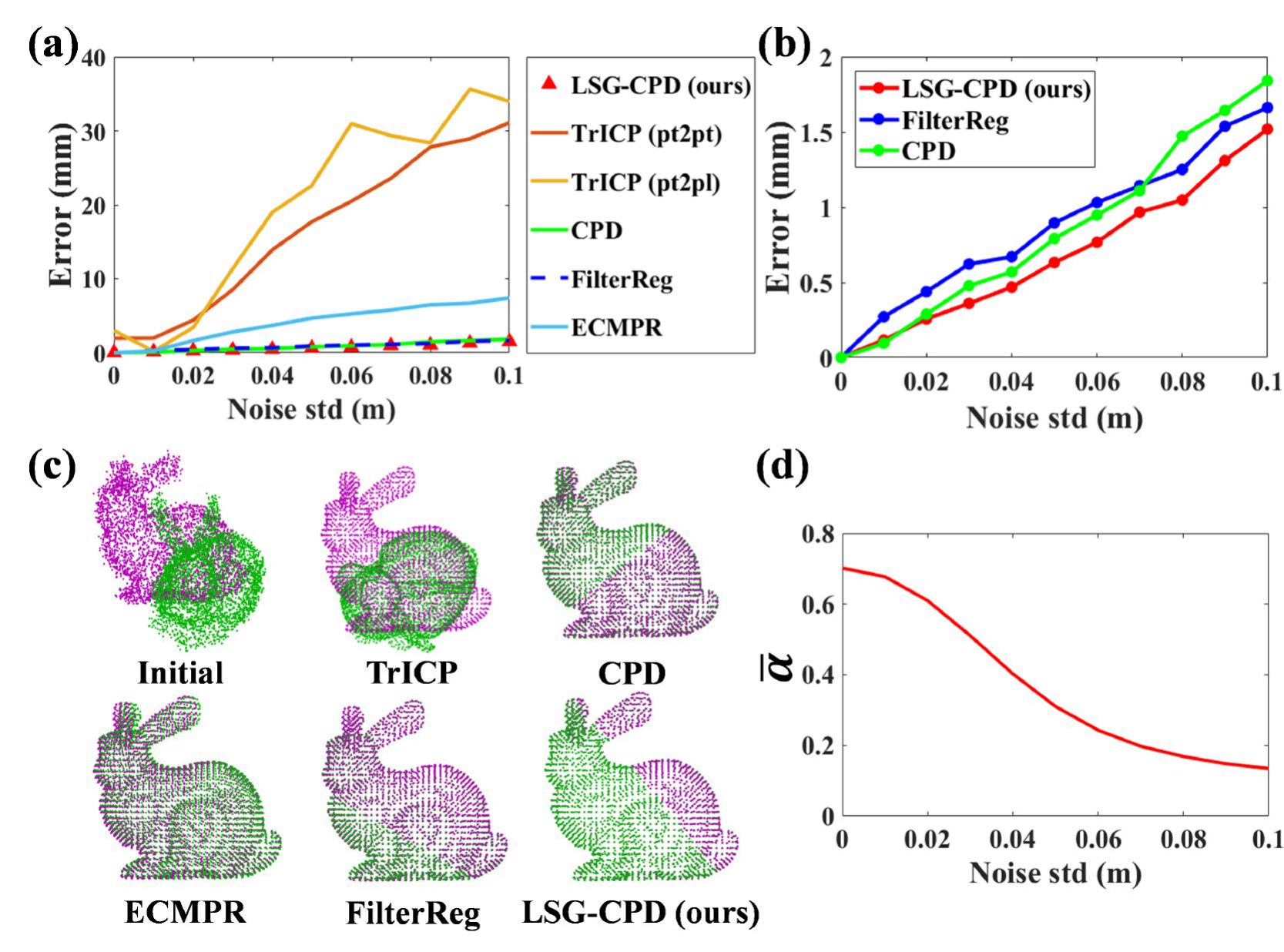} 
    \caption{Noise results. (a) Errors (Eq. \eqref{eqn: multiview error}) of different methods. (b) Enlarged image showing errors of CPD, FilterReg, and LSG-CPD. (c) Initialization and results of different methods on data with 0.03m noise level. TrICP point-to-plane result is shown as the result for TrICP. Denoised point clouds are used for better visualization of registration result. (d) Averaged penalization coefficient $\bar{\alpha}$ (Eq. \eqref{eqn:2}) of all points in the target set with respect to the noise level.}
    \label{fig:noise}
    \vspace{-0.4cm}
\end{figure}

In the outlier experiment, we add different ratios (the number of outliers to the number of original points) of Gaussian outliers. 
Results are shown in Fig. \ref{fig:outlier}. 
All the GMM-based methods significantly outperform TrICP when there are outliers present, where CPD, FilterReg, and the proposed method are the best.
However, with the increase of the outlier ratio, the error of CPD and FilterReg increases while the performance of the proposed method maintains and outperforms CPD and FilterReg (Fig. \ref{fig:outlier}(b)).

\begin{figure*}[!htp]
    \centering
    \includegraphics[scale=0.46]{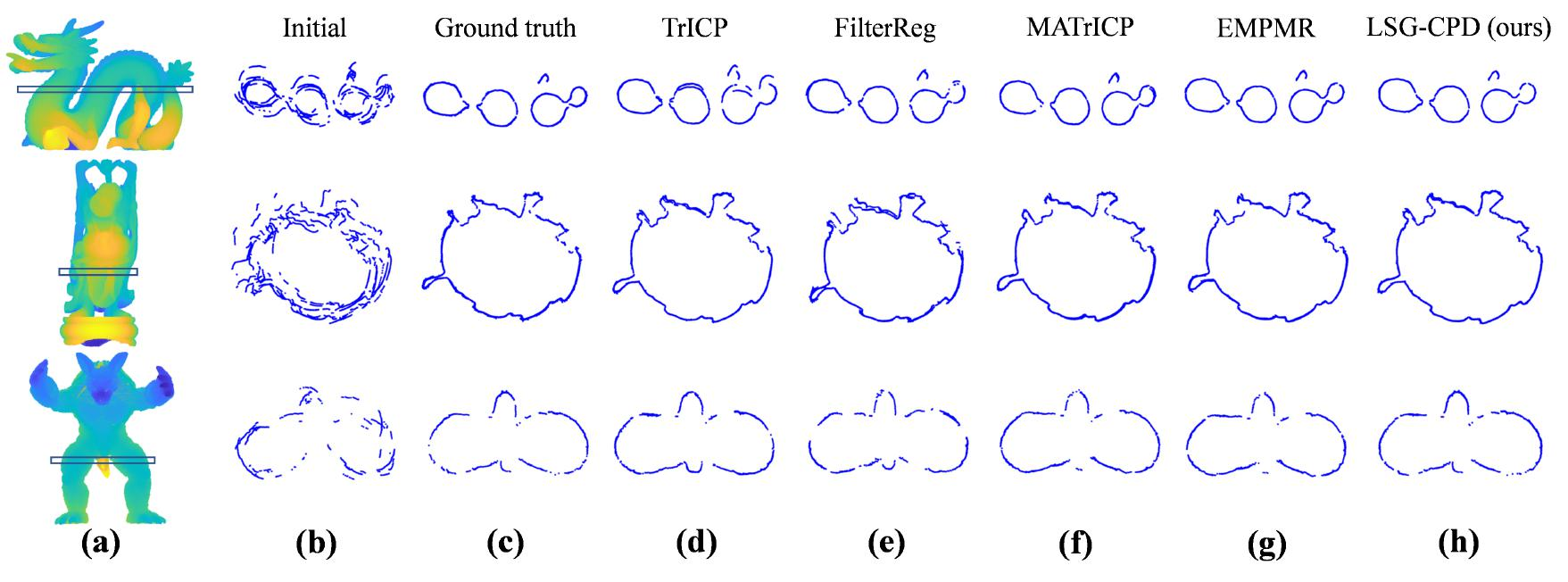}
    \caption{Multiview cross-section results where the corresponding regions are indicated in (a). Thinner curves indicate better registration.  (a) shows the whole object modeling results of the proposed method; (b) and (c) show the initial and ground truth; the rest show the cross-section results of different methods.}
    \label{fig:multiview}
    \vspace{-0.4cm}
\end{figure*}

In the noise experiment, we corrupt each point in the source and target set by adding Gaussian noise with different std.
Results are shown in Fig. \ref{fig:noise}.
All the GMM-based methods significantly outperform TrICP when there is noise.
CPD, FilterReg, and the proposed methods have similar performance among which the proposed method has a small advantage (Fig. \ref{fig:noise}(b)).
We also show the averaged penalization coefficient $\bar{\alpha}$ at different noise levels (Fig. \ref{fig:noise}(d)). 
As the noise level increases, $\bar{\alpha}$ decreases.
This is because without noise, the local surface will look like a plane and thus $\alpha$ will be relatively large; when the points are corrupted, the local plane will also be corrupted and normal estimation becomes less effective, resulting in small $\alpha$ as we desire.

We compare the speed with CPD on data with different outlier ratios.
In this experiment, we terminate the algorithm when the rotation error is less than $8\degree$ and the translation error is less than 1cm.
Fig. \ref{fig:outlier}(d) and (e) summarize the result.
Our method is 5 times faster than CPD (89ms vs 450ms) when there are no outliers and 29 times faster (346ms vs 9940ms) when the outlier ratio is 1.
Generally, our method takes less iterations to converge.
And the time for each iteration is less than that of CPD.
Averaging across all the experiment, our method is 20 times faster than CPD.

For FilterReg, we use the point-to-point version.
We tested the point-to-plane version extensively but find it doesn't give a good performance.
We think the reasons could be the singular covariance and the inconsistency of the probabilistic model in the E and M step, as mentioned in Sec. \ref{sec: math}, making it not robust to bad initialization and/or outliers and noise.
In the following experiments, if not mentioned, FilterReg is tested with the point-to-point version.

\begin{table}
\begin{center}
\begin{tabular}{c c c c}
\toprule
 Method & Dragon & Happy & Armadillo \\
 \arrayrulecolor{black}\midrule
Initial     & 5.12 & 5.41 & 5.23 \\
TrICP       & 1.87 & 1.16 & 0.66 \\
FilterReg   & 1.13 & 2.13 & 0.63 \\
MATrICP     & 1.18 & 1.29 & 0.53 \\
EMPMR       & 1.39 & \textbf{1.07} & 0.50 \\
 \arrayrulecolor{black!30}\midrule
LSG-CPD (ours) & \textbf{0.97} & 1.08 & \textbf{0.43} \\
 \arrayrulecolor{black}\bottomrule
\end{tabular}
\end{center}
\caption{Object Modeling Errors (Eq. \eqref{eqn: multiview error}, unit mm).}
\label{table: multivew}
\vspace{-0.6cm} 
\end{table}

\subsection{Object Modeling on Range Datasets}
\label{sec: object modeling}
In this experiment, we test on object modeling from laser range data of objects captured from different viewpoints.
We evaluate on three datasets from the Stanford 3D scanning repository \cite{curless1996volumetric}, \textit{i.e.}, Dragon Stand, Happy Stand, and Armadillo.
Each consisting of 12 or 15 sequential scans of an object with evenly spaced view angles.
We compare our method against two pairwise methods, \textit{i.e.}, FilterReg \cite{gao2019filterreg} and TrICP \cite{chetverikov2002trimmed}, and two multi-view methods, \textit{i.e.}, MATrICP \cite{li2014improved} and EMPMR \cite{zhu2020registration}.
Object modeling for pairwise methods makes use of the spatial information provided by the scanning sequence.
Point clouds that are spatially adjacent to each other are first pairwisely registered. 
Then, all the point clouds are aligned to the first one for whole object modeling.
Following \cite{zhu2020registration}, we perturb the point clouds from its ground truth pose with a random pose ($\pm0.05$rad in rotation and $\pm0.01$m in translation).
We downsample the point cloud of each scan with voxelized grids, resulting in about 3500 points for each view.
For each dataset, we carry out 20 independent experiments.
We use the averaged corresponding point-to-point distance (Eq. \eqref{eqn: multiview error}) as the evaluation metric.

Qualitative results are shown in Table \ref{table: multivew}.
Fig. \ref{fig:multiview} shows the cross-section results of different methods.
The proposed method outperforms the two pairwise baseline methods.
The chain-like reconstruction method used for pairwise methods has a disadvantage of error accumulation.
The error in each pairwise registration will propagate and accumulate along the chain when aligned to the base point cloud.
On the other hand, multi-view methods can simultaneously process all the point clouds and take advantage of techniques such as motion averaging \cite{li2014improved}.
Even with such a significant disadvantage, compared with state-of-the-art multi-view methods, our method outperforms them on two datasets and is on par with them on the third.

\begin{figure}
    \centering
    \includegraphics[width=0.95\columnwidth]{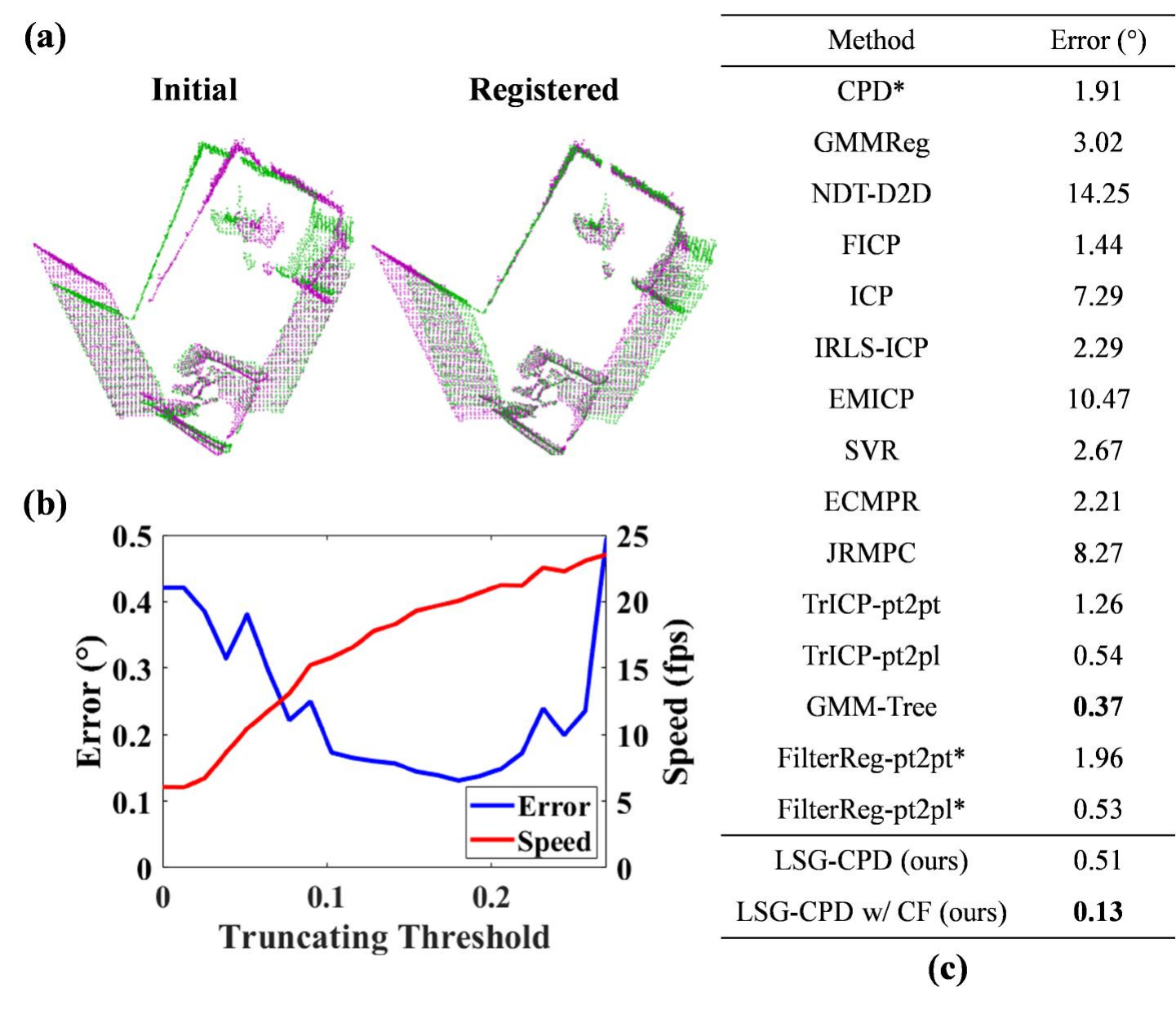}
    \caption{Stanford Lounge Result. (a) Examples of registration result. (b) Angle error and speed versus truncating threshold. (c) Registration results of various methods. The results of the baseline methods with a star (*) are sourced from \cite{gao2019filterreg}; the results of other baseline methods are sourced from \cite{eckart2018hgmr}.}
    \label{fig:lounge}
    \vspace{-0.4cm}
\end{figure}

\subsection{Registration on RGBD Dataset}
\label{sec:RGBD}
In this section, we follow the setting in \cite{eckart2018hgmr} and test on the Stanford Lounge dataset \cite{zhou2013dense} which contains range data of an indoor scene captured with an RGBD camera.
We register every 5-th frame for the first 400 frames. 
Each frame is downsampled to about 5000 points.
Following \cite{eckart2018hgmr}, we use the average Euler angle deviation from the ground truth to measure the rotation error between each pair.

The results are shown in Fig. \ref{fig:lounge}.
Our method achieves the state-of-the-art without CF. 
With CF, the accuracy improves greatly and outperforms the state-of-the-art. 
The angle error is more than 2 times lower than the state-of-the-art. 
Additionally, CF greatly accelerates the registration by truncating points with lower accuracy confidence. 
The registration frame rate increases from 5 FPS to 19 FPS.
\begin{figure*}[!htp]
    \centering
    \includegraphics[scale=0.46]{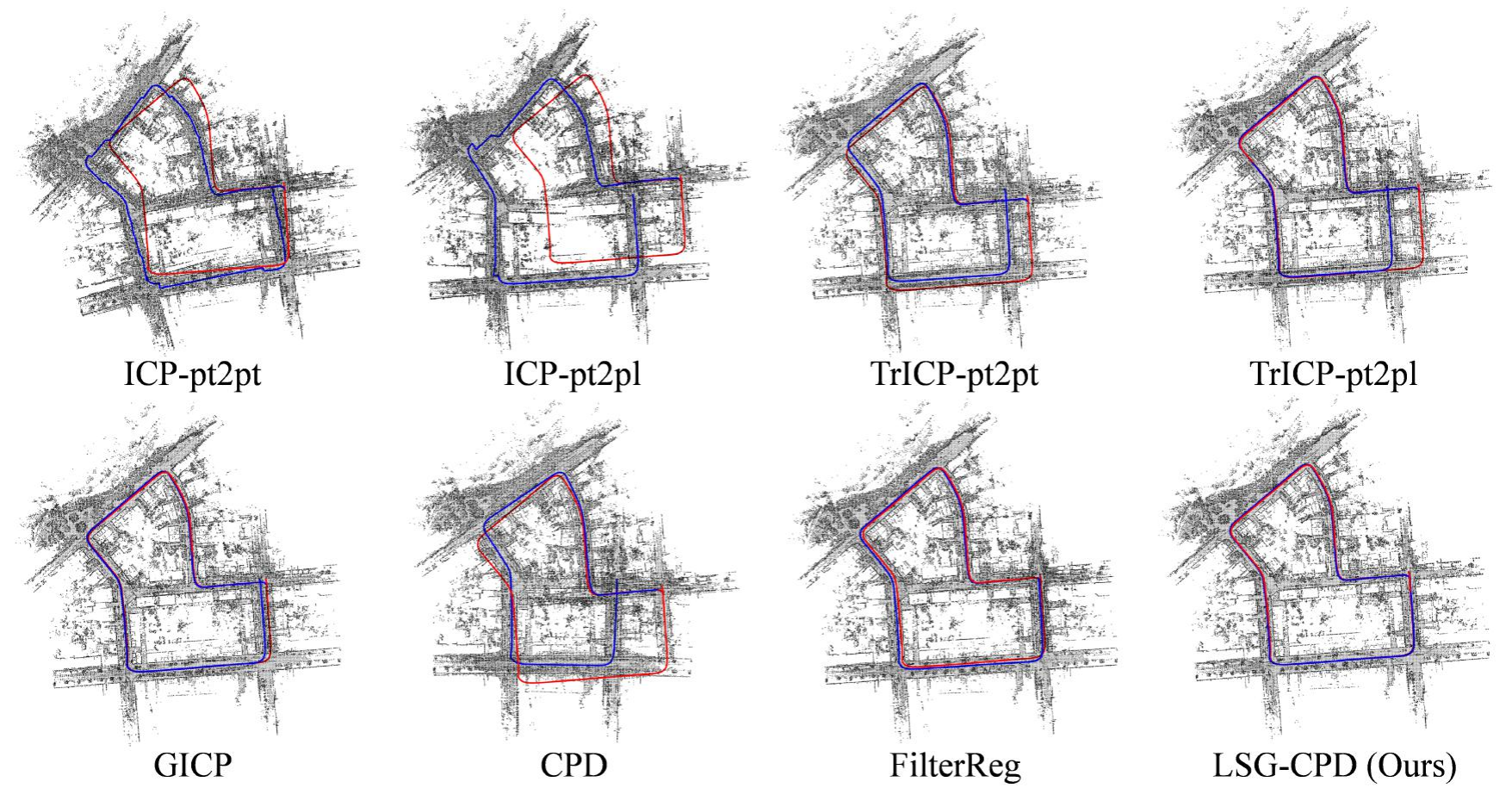} 
    \caption{Kitti Odometry Benchmark 07 Sequence Result Visualization. Red indicates the ground truth; blue indicates the registration result.}
    \label{fig:kitti}
\end{figure*}
\begin{table*}[!h]
\begin{center}
\begin{tabular}{c c c c c c c c} 
\toprule
Method & Rel. Rot. (\degree) & Rel. Tran. (m) & Abs. Rot. (\degree) & Abs. Tran. (m) & Last Rot. (\degree) & Last Tran. (m)  \\
\arrayrulecolor{black}\midrule
\rowcolor{lightgray}ICP-pt2pt & 0.283 (3.690) & 0.214 (5.966) & 9.476 (17.641) & 17.296 (36.379) & 7.100 & 16.131\\
ICP-pt2pl & 0.475 (8.683) & 0.187 (14.320) & 7.986 (16.496) & 48.422 (70.712) & 6.526 & 58.902\\
\rowcolor{lightgray}TrICP-pt2pt & 0.123 (0.689) & 0.117 (2.324) & 1.606 (2.476) & 10.446 (24.881) & 1.587 & 24.449\\
TrICP-pt2pl & 0.071 (0.607) & 0.080 (2.604) & 1.365 (3.150) & 8.914 (33.113) & 3.073 & 33.094\\
\rowcolor{lightgray}GICP & 0.070 (0.787) & 0.035 (2.108) & 1.181 (1.799) & 3.052 (6.939) & 1.772 & 6.593\\
CPD & 0.167 (1.100) & 0.190 (2.701) & 5.091 (7.900) & 21.696 (56.690) & 6.353 & 47.359\\
\rowcolor{lightgray}FilterReg & 0.091 (0.615) & 0.029 (0.099) & 1.995 (7.016) & 3.812 (5.666) &6.984 & 5.373\\
LSG-CPD (ours)  & \textbf{0.062 (0.591)} & \textbf{0.022 (0.090)} & \textbf{0.812 (1.419)} & \textbf{1.069 (1.830)} & \textbf{1.377} & \textbf{0.382} \\
\arrayrulecolor{black}\bottomrule
\end{tabular}
\end{center}
\caption{Kitti Odometry Benchmark 07 Sequence Result. We show the relative rotation error (Rel. Rot.), relative translation error (Rel. Tran.), absolute rotation error (Abs. Rot.), absolute translation error (Abs. Tran.), last frame rotation error (Last Rot.), and last frame translation error (Last Tran.) of different methods. In the first four columns of result, the value outside the parenthesis is the averaged error of the 550 pairs; the value inside is the maximum.}
\label{table: kitti}
\vspace{-0.4cm}
\end{table*}
We also experiment on truncating points at different confidence thresholds (Fig. \ref{fig:lounge}(b)).
The registration error decreases from $0.51\degree$ to $0.42\degree$ by just adding CF without any truncation. 
As expected, the registration speed increases as the truncating threshold increases.
However, if the threshold is too large, too many points, including the accurately measured points, are truncated. 
The point cloud will contain insufficient information for registration.
If the threshold is too small, \textit{i.e.}, too few points are truncated, the point cloud contains many noisy data. 

\subsection{Registration on LiDAR Dataset}
\label{sec:lidar}
In this experiment, we evaluate on the Kitti dataset \cite{geiger2012we}.
We test on the 07 sequence of the odometry benchmark, which contains a drive for about 700 meters.
We downsample the point clouds to about 5000 points with voxelized grids.
We register together every other frame pairwisely for the whole sequence, resulting in 550 pairs of point clouds for registration.
All the frames are aligned to the first frame to reconstruct the whole sequence.
We evaluate the averaged relative and absolute rotation and translation errors.
We also compute the rotation and translation errors of the last frame.
The translation error is the Euclidean distance between the registration result and the ground truth.
The rotation error $\Delta \theta$ is the angle of the axis-angle parameterization of the relative rotation between the ground truth and the registration result.
The results are shown in Fig. \ref{fig:kitti} and Table \ref{table: kitti}.
Our method outperforms all baseline methods on every evaluation metric.
Remarkably, even without loop closure, our proposed method is able to achieve very small translation and rotation errors in the last frame of the whole sequence.
All the ICP-based methods except GICP \cite{Segal-RSS-09} have relatively large errors.
For FilterReg\cite{gao2019filterreg}, we use the point-to-plane version as the point-to-point version does not give a good result.
FilterReg \cite{gao2019filterreg} and the proposed method outperform CPD \cite{myronenko2010point}.

\section{Discussion \& Future Works}
Extensive experiments on various kinds of data show that our method is accurate and robust.
As shown in Sec. \ref{sec: Outlier and Noise}, ICP-based methods require good initialization and are not robust to outliers and noise.
CPD and point-to-point FilterReg, although robust to outliers, noise, and missing points, do not perform well on the Kitti dataset.
We hypothesize this is because LiDAR data contain many large planes.
The isotropic covariance in CPD and point-to-point FilterReg are not able to accurately represent the geometry of the data, thus leading to bad performance.
Point-to-plane FilterReg performs well on the Stanford Lounge and Kitti dataset in which the relative pose between adjacent frames for registration are close to each other.
However, it is sensitive to bad initialization and are not robust to outliers and noise as discussed in Sec. \ref{sec: math} and \ref{sec: Outlier and Noise}.
Our method is robust to outliers, noise, and missing points and accurate.
The reason for the robustness is twofold.
The probabilistic modeling which includes the outlier component make it robust to outliers and missing points like other GMM-based methods \cite{myronenko2010point, gao2019filterreg}.
The covariance adaptively adjusts its shape according to the local surface geometry, making it insusceptible to ineffectively estimated surface normals.
The reason for the accuracy is also credit to the local surface geometry-based covariance.
By using the surface variation \cite{surface_variance} to adjust the point-to-plane penalization in the covariance, our method can accurately represent the geometry of a wide variety of data, including data of different scales, with/without large planes.
Future works include extending the current framework to non-rigid registration and applying the method to more computer vision tasks, \textit{e.g.}, pose estimation and SLAM.

\section{Conclusions}
We present a novel GMM-based probabilistic registration method which outperforms state-of-the-art algorithms in terms of accuracy and robustness on various datasets.
Our method makes use of the surface normal and surface variation \cite{surface_variance} to create anisotropic covariances in the GMM which faithfully represents the geometry of the point cloud.
We show that the computation can be reformulated into simple matrix manipulations which can be efficiently computed on a GPU.
In addition, we propose an efficient optimization method to solve for the optimal transformation on a matrix Lie group in the EM process.
Thanks to the efficient computation and optimization, the proposed method is about 20 times faster than modern implementations of CPD \cite{myronenko2010point}.

\vspace{\baselineskip}
\noindent \textbf{Acknowledgements}
This work was supported by NUS Startup grants R-265-000-665-133, R-265-000-665-731, Faculty Board account C-265-000-071-001, MOE grant R-265-000-655-114, and JHU internal funds.

{\small
\bibliographystyle{ieee_fullname}
\bibliography{egbib}

\begin{thebibliography}{10}\itemsep=-1pt

\bibitem{basso2018robust}
F. Basso, E. Menegatti, and A. Pretto.
\newblock Robust intrinsic and extrinsic calibration of rgb-d cameras.
\newblock {\em IEEE Transactions on Robotics}, 34(5):1315--1332, 2018.

\bibitem{besl1992ICP}
P.~J. Besl and N.D. McKay.
\newblock A method for registration of 3-d shapes.
\newblock {\em IEEE transactions on pattern analysis and machine intelligence},
  14(2):239--256, 1992.

\bibitem{10.1007/978-3-319-10404-1_23}
S. Billings and R. Taylor.
\newblock Iterative most likely oriented point registration.
\newblock In Polina Golland, Nobuhiko Hata, Christian Barillot, Joachim
  Hornegger, and Robert Howe, editors, {\em Medical Image Computing and
  Computer-Assisted Intervention -- MICCAI 2014}, pages 178--185, Cham, 2014.
  Springer International Publishing.

\bibitem{bishop2006pattern}
C.~M. Bishop.
\newblock {\em Pattern recognition and machine learning}.
\newblock springer, 2006.

\bibitem{briales2017convex}
Jesus Briales and Javier Gonzalez-Jimenez.
\newblock Convex global 3d registration with lagrangian duality.
\newblock In {\em Proceedings of the IEEE Conference on Computer Vision and
  Pattern Recognition}, 2017.

\bibitem{campbell2015adaptive}
D. Campbell and L. Petersson.
\newblock An adaptive data representation for robust point-set registration and
  merging.
\newblock In {\em Proceedings of the IEEE International Conference on Computer
  Vision}, pages 4292--4300, 2015.

\bibitem{chen1992object}
Y. Chen and G. Medioni.
\newblock Object modelling by registration of multiple range images.
\newblock {\em Image and vision computing}, 10(3):145--155, 1992.

\bibitem{chetverikov2002trimmed}
D. Chetverikov, D. Svirko, D. Stepanov, and P. Krsek.
\newblock The trimmed iterative closest point algorithm.
\newblock In {\em Pattern Recognition, International Conference on}, volume~3,
  pages 30545--30545, 2002.

\bibitem{stochastic}
G.~S. Chirikjian.
\newblock {\em Stochastic Models, Information Theory, and Lie Groups, Volume
  2}.
\newblock 01 2012.

\bibitem{christian2013survey}
J.~A. Christian and S. Cryan.
\newblock A survey of lidar technology and its use in spacecraft relative
  navigation.
\newblock In {\em AIAA Guidance, Navigation, and Control (GNC) Conference},
  page 4641, 2013.

\bibitem{curless1996volumetric}
B. Curless and M. Levoy.
\newblock A volumetric method for building complex models from range images.
\newblock In {\em Proceedings of the 23rd annual conference on Computer
  graphics and interactive techniques}, pages 303--312, 1996.

\bibitem{DEMP1977}
A.~P. Dempster, N.~M. Laird, and D.~B. Rubin.
\newblock Maximum likelihood from incomplete data via the {EM} algorithm.
\newblock {\em Journal of the Royal Statistical Society: Series B}, 39:1--38,
  1977.

\bibitem{eckart2018hgmr}
B. Eckart, K. Kim, and J. Kautz.
\newblock Hgmr: Hierarchical gaussian mixtures for adaptive 3d registration.
\newblock In {\em Proceedings of the European Conference on Computer Vision
  (ECCV)}, pages 705--721, 2018.

\bibitem{gao2019filterreg}
W. Gao and R. Tedrake.
\newblock Filterreg: Robust and efficient probabilistic point-set registration
  using gaussian filter and twist parameterization.
\newblock In {\em Proceedings of the IEEE Conference on Computer Vision and
  Pattern Recognition}, pages 11095--11104, 2019.

\bibitem{geiger2012we}
A. Geiger, P. Lenz, and R. Urtasun.
\newblock Are we ready for autonomous driving? the kitti vision benchmark
  suite.
\newblock In {\em 2012 IEEE Conference on Computer Vision and Pattern
  Recognition}, pages 3354--3361. IEEE, 2012.

\bibitem{granger2002multi}
S. Granger and X. Pennec.
\newblock Multi-scale em-icp: A fast and robust approach for surface
  registration.
\newblock In {\em European Conference on Computer Vision}, pages 418--432.
  Springer, 2002.

\bibitem{depth_camera_error_model}
G. {Halmetschlager-Funek}, M. {Suchi}, M. {Kampel}, and M. {Vincze}.
\newblock An empirical evaluation of ten depth cameras: Bias, precision,
  lateral noise, different lighting conditions and materials, and multiple
  sensor setups in indoor environments.
\newblock {\em IEEE Robotics Automation Magazine}, 26(1):67--77, 2019.

\bibitem{hirose2020bayesian}
O. Hirose.
\newblock A bayesian formulation of coherent point drift.
\newblock {\em IEEE transactions on pattern analysis and machine intelligence},
  2020.

\bibitem{hoppe1992surface}
H. Hoppe, T. DeRose, T. Duchamp, J. McDonald, and W. Stuetzle.
\newblock Surface reconstruction from unorganized points.
\newblock In {\em Proceedings of the 19th annual conference on computer
  graphics and interactive techniques}, pages 71--78, 1992.

\bibitem{horaud2010rigid}
R. Horaud, F. Forbes, M. Yguel, G. Dewaele, and J. Zhang.
\newblock Rigid and articulated point registration with expectation conditional
  maximization.
\newblock {\em IEEE Transactions on Pattern Analysis and Machine Intelligence},
  33(3):587--602, 2010.

\bibitem{horaud2016overview}
R. Horaud, M. Hansard, G. Evangelidis, and C. M{\'e}nier.
\newblock An overview of depth cameras and range scanners based on
  time-of-flight technologies.
\newblock {\em Machine vision and applications}, 27(7):1005--1020, 2016.

\bibitem{jian2010robust}
B. Jian and B.~C. Vemuri.
\newblock Robust point set registration using gaussian mixture models.
\newblock {\em IEEE transactions on pattern analysis and machine intelligence},
  33(8):1633--1645, 2010.

\bibitem{li2014improved}
Z. Li, J. Zhu, K. Lan, C. Li, and C. Fang.
\newblock Improved techniques for multi-view registration with motion
  averaging.
\newblock In {\em 2014 2nd International Conference on 3D Vision}, volume~1,
  pages 713--719. IEEE, 2014.

\bibitem{luo2003unified}
B. Luo and E.~R. Hancock.
\newblock A unified framework for alignment and correspondence.
\newblock {\em Computer Vision and Image Understanding}, 92(1):26--55, 2003.

\bibitem{mcneill2006probabilistic}
G. McNeill and S. Vijayakumar.
\newblock A probabilistic approach to robust shape matching.
\newblock In {\em 2006 International Conference on Image Processing}, pages
  937--940. IEEE, 2006.

\bibitem{min2019_nonrigid}
Z. Min, J. Pan, A. Zhang, and M.~Q.-H. Meng.
\newblock Robust non-rigid point set registration algorithm considering
  anisotropic uncertainties based on coherent point drift.
\newblock In {\em IEEE/RSJ International Conference on Intelligent Robots and
  Systems}, 2019.

\bibitem{min2019robust}
Z. Min, J. Wang, and M.~Q.-H. Meng.
\newblock Robust generalized point cloud registration with orientational data
  based on expectation maximization.
\newblock {\em IEEE Transactions on Automation Science and Engineering},
  17(1):207--221, 2019.

\bibitem{murray1994mathematical}
R.~M. Murray, Z. Li, and S.~S. Sastry.
\newblock {\em A mathematical introduction to robotic manipulation}.
\newblock CRC press, 1994.

\bibitem{myronenko2010point}
A. Myronenko and X. Song.
\newblock Point set registration: Coherent point drift.
\newblock {\em IEEE transactions on pattern analysis and machine intelligence},
  32(12):2262--2275, 2010.

\bibitem{nuchter20076d}
A. N{\"u}chter, K. Lingemann, J. Hertzberg, and H. Surmann.
\newblock 6d slam—3d mapping outdoor environments.
\newblock {\em Journal of Field Robotics}, 24(8-9):699--722, 2007.

\bibitem{surface_variance}
M. {Pauly}, M. {Gross}, and L.~P. {Kobbelt}.
\newblock Efficient simplification of point-sampled surfaces.
\newblock In {\em IEEE Visualization, 2002. VIS 2002.}, pages 163--170, 2002.

\bibitem{rangarajan1997robust}
A. Rangarajan, H. Chui, E. Mjolsness, S. Pappu, L. Davachi, P. Goldman-Rakic,
  and J. Duncan.
\newblock A robust point-matching algorithm for autoradiograph alignment.
\newblock {\em Medical image analysis}, 1(4):379--398, 1997.

\bibitem{ravikumar2017generalised}
N. Ravikumar, A. Gooya, A.~F. Frangi, and Z.~A. Taylor.
\newblock Generalised coherent point drift for group-wise registration of
  multi-dimensional point sets.
\newblock In {\em International Conference on Medical Image Computing and
  Computer-Assisted Intervention}, pages 309--316. Springer, 2017.

\bibitem{schmidt2014dart}
T. Schmidt, R.~A. Newcombe, and D. Fox.
\newblock Dart: Dense articulated real-time tracking.
\newblock In {\em Proceedings of Robotics: Science and Systems}, 2014.

\bibitem{Segal-RSS-09}
A. Segal, D. Haehnel, and S. Thrun.
\newblock Generalized-icp.
\newblock In {\em Proceedings of Robotics: Science and Systems}, Seattle, USA,
  June 2009.

\bibitem{NICP}
J. {Serafin} and G. {Grisetti}.
\newblock Nicp: Dense normal based point cloud registration.
\newblock In {\em 2015 IEEE/RSJ International Conference on Intelligent Robots
  and Systems (IROS)}, pages 742--749, 2015.

\bibitem{stoyanov2012fast}
T. Stoyanov, M. Magnusson, H. Andreasson, and A.~J. Lilienthal.
\newblock Fast and accurate scan registration through minimization of the
  distance between compact 3d ndt representations.
\newblock {\em The International Journal of Robotics Research},
  31(12):1377--1393, 2012.

\bibitem{tabib2018manifold}
W. Tabib, C. O’Meadhra, and N. Michael.
\newblock On-manifold gmm registration.
\newblock {\em IEEE Robotics and Automation Letters}, 3(4):3805--3812, 2018.

\bibitem{whelan2015elasticfusion}
T. Whelan, S. Leutenegger, R. Salas-Moreno, B. Glocker, and A. Davison.
\newblock Elasticfusion: Dense slam without a pose graph.
\newblock In {\em Proceedings of Robotics: Science and Systems}, 2015.

\bibitem{wu1983EMConverge}
C.~F. Wu.
\newblock On the convergence properties of the em algorithm.
\newblock {\em Ann. Statist.}, 11(1):95--103, 03 1983.

\bibitem{yang2020teaser}
H. Yang, J. Shi, and L. Carlone.
\newblock Teaser: Fast and certifiable point cloud registration.
\newblock {\em IEEE Transactions on Robotics}, 37(2):314--333, 2021.

\bibitem{zhou2013dense}
Q.-Y. Zhou and V. Koltun.
\newblock Dense scene reconstruction with points of interest.
\newblock {\em ACM Transactions on Graphics (ToG)}, 32(4):1--8, 2013.

\bibitem{zhu2020registration}
J. Zhu, R. Guo, Z. Li, J. Zhang, and S. Pang.
\newblock Registration of multi-view point sets under the perspective of
  expectation-maximization.
\newblock {\em IEEE Transactions on Image Processing}, 29:9176--9189, 2020.

\end{thebibliography}
}

\end{document}